%% file: main.tex
\documentclass[10pt,twocolumn,letterpaper]{article}

\usepackage[pagenumbers]{cvpr} 

\input{preamble}

\definecolor{cvprblue}{rgb}{0.21,0.49,0.74}
\usepackage[pagebackref,breaklinks,colorlinks,allcolors=cvprblue]{hyperref}

\def\paperID{} 
\def\confName{CVPR}
\def\confYear{2026}

\title{Learning 3D Reconstruction with Priors in Test Time}

\author{
Lei Zhou\thanks{Email: \texttt{lezzhou@cs.stonybrook.edu}} \quad Haoyu Wu \quad Akshat Dave \quad Dimitris Samaras\\
Stony Brook University
}

\begin{document}
\maketitle
\input{sec/0_abstract}    
\input{sec/1_intro}
\input{sec/2_relatedwork}
\input{sec/3_method}
\input{sec/4_experiments}
\input{sec/5_conclusion}
\clearpage
\input{sec/6_ack}
{
    \small
    \bibliographystyle{ieeenat_fullname}
    \bibliography{main}
}
\input{sec/X_suppl}

\end{document}

%% file: preamble.tex
\usepackage{algorithm}
\usepackage{algorithmic}

\usepackage{graphicx}
\usepackage{booktabs}
\usepackage{multirow}

\newcommand{\tablestyle}[2]{%
  \setlength{\tabcolsep}{#1}%
  \renewcommand{\arraystretch}{#2}%
}

\usepackage{pifont}
\newcommand{\xmark}{\ding{55}}

%% file: sec/0_abstract.tex
\begin{abstract}
    We introduce a test-time framework for multiview Transformers (MVTs) that
    incorporates priors (e.g., camera poses, intrinsics, and depth) to improve 3D tasks
    without retraining or modifying pre-trained image-only networks.
    Rather than feeding priors into the architecture,
    we cast them as constraints on the predictions and optimize the network at inference time.
    The optimization loss consists of a self-supervised objective and prior penalty terms.
    The self-supervised objective captures the compatibility among multi-view predictions
    and is implemented using photometric or geometric loss between renderings from other views and each view itself.
    Any available priors are converted into penalty terms on the corresponding output modalities.
    Across a series of 3D vision benchmarks, including point map estimation and camera pose estimation,
    our method consistently improves performance over base MVTs by a large margin.
    On the ETH3D, 7-Scenes, and NRGBD datasets, our method reduces the point-map distance error by more than half compared with the base image-only models.
    Our method also outperforms retrained prior-aware feed-forward methods,
    demonstrating the effectiveness of our test-time constrained optimization (TCO) framework for incorporating priors into 3D vision tasks.
    The code is available at \url{https://github.com/cvlab-stonybrook/TCO}.
\end{abstract}

%% file: sec/1_intro.tex
\section{Introduction}
\label{sec:intro}
Reconstructing 3D geometry from images is a fundamental problem in computer vision,
with potential applications in autonomous driving~\cite{elluswamy2025teslaai}, 
augmented reality~\cite{engel2023project}, 
and robotics~\cite{qian20243d}.
Traditional methods, such as Structure from Motion (SfM)
~\cite{hartley2003multiple, szeliski2022computer, schonberger2016structure, pan2024global},
approach the problem with multi-stage pipelines,
including feature matching~\cite{lowe2004distinctive, detone2018superpoint, edstedt2024roma, sun2021loftr, sarlin2020superglue, lindenberger2023lightglue}
and bundle adjustment~\cite{triggs1999bundle, schonberger2016structure},
that are optimized separately.
To address this,
feed-forward multiview transformers (MVTs)~\cite{wang2023dust3r,leroy2024grounding, cabon2025must3r, wang2025vggt, wang2025pi, keetha2025mapanything, jang2025pow3r} are proposed to perform 3D reconstruction from a set of images with just one forward pass.
Thanks to large-scale end-to-end training,
MVTs achieve impressive results.

However,
MVTs~\cite{wang2025vggt, wang2023dust3r, leroy2024grounding, cabon2025must3r, wang2025pi}
are natively designed to only accept RGB images as input.
This limits their applications in real-world scenarios where extra information is available.
For example, camera poses can be obtained from a camera rig system~\cite{jensen2014large}
or sophisticated SfM methods, such as COLMAP~\cite{schonberger2016structure},
and can be used to improve geometry reconstruction quality.
In another scenario, depth maps can be obtained from depth sensors, such as LiDAR or stereo cameras,
which complement RGB images for improving camera pose estimation~\cite{zheng2022fast, zheng2024fast}.

More recent works~\cite{jang2025pow3r, keetha2025mapanything} modify MVTs to accept additional priors as input,
including camera poses and depth maps.
However, these methods are bound to specific architectures and types of priors.
They must be retrained whenever the backbone or the available priors change,
which is inflexible and computationally expensive.

In this paper,
we propose a simple test-time constrained optimization (TCO) framework that incorporates priors without changing or retraining the original MVTs.
Rather than feeding priors as input,
we treat them as constraints on the model’s outputs and optimize the network to fit them at inference.
The idea is to leverage the ground truths (e.g., priors) of some prediction modalities to improve the other prediction modalities as well.
To prevent the optimization from overfitting to the priors without improving the overall 3D task quality,
the optimization loss consists of prior penalty terms and a self-supervised objective.
The self-supervised objective is defined as the compatibility among MVTs' predictions,
implemented by the photometric or geometric loss between the renderings projected from other views and each view itself.
To exploit the synergy between the different prediction modalities,
we adopt a fine-tuning strategy that freezes all the prediction heads but fine-tunes only the shared decoder network with LoRA~\cite{hu2022lora}.

We evaluate our method across standard 3D vision benchmarks,
including point-map estimation with camera poses and intrinsics
and camera pose estimation with depth maps.
We observe consistent improvements over base MVTs across all the benchmarks.
On ETH3D~\cite{schops2017multi}, 7-Scenes~\cite{shotton2013scene}, and NRGBD~\cite{azinovic2022neural},
we reduce the error by more than half compared with the base image-only models.
When compared with other prior-aware methods that retrain a new model by feeding priors as input,
our method also shows superior performance.
These results prove the effectiveness of our TCO framework in incorporating priors for 3D tasks.
In summary, our contributions are:
\begin{itemize}
    \item A test-time constrained optimization (TCO) framework that incorporates scene and camera priors without retraining or architectural changes.
    \item A TCO loss design that includes a self-supervised objective to enforce multi-view prediction compatibility and prior penalty terms.
    \item Extensive evaluations on 3D reconstruction benchmarks showing consistent large-margin improvements over base MVTs, surpassing prior-aware feed-forward methods while remaining plug-and-play.
\end{itemize}

%% file: sec/2_relatedwork.tex
\section{Related Work}
\label{sec:relatedwork}
\paragraph{Feed-forward 3D Reconstruction (No Priors).}
Reconstructing scenes from images has long relied on classical Structure-from-Motion (SfM) pipelines
that estimate camera parameters and triangulate structure through multi-stage optimization \cite{schonberger2016structure, pan2024global, hartley2003multiple, szeliski2022computer}.
Recent feed-forward models bypass iterative optimization by directly regressing 3D structure from one or more views.
DUSt3R predicts pairwise pointmaps anchored to a reference frame and recovers relative poses post hoc \cite{wang2023dust3r},
while MASt3R augments the representation with dense features for robust matching \cite{leroy2024grounding}.
Multi-view variants reduce pairwise redundancy:
MUSt3R and VGGT process many images jointly \cite{cabon2025must3r, wang2025vggt},
and $\pi^3$ further improves scalability with permutation-equivariant training that avoids fixing a reference frame \cite{wang2025pi}.

\paragraph{Feed-forward 3D Reconstruction with Priors.}
Several feed-forward systems optionally leverage geometric priors at inference to improve 3D reconstruction or enable metric predictions.
Such guidance includes intrinsics/extrinsics, sparse 3D points, or monocular depth priors.
Pow3R demonstrates that conditioning on any subset of camera parameters and sparse points improves two-view pointmap prediction \cite{jang2025pow3r}.
MapAnything supports flexible inputs, seamlessly incorporating priors to predict metric geometry \cite{keetha2025mapanything}.
Related lines of work show how intrinsics can enhance monocular depth estimation (e.g., UniDepth~\cite{piccinelli2024unidepth}).
Geometry, e.g., depth maps, has also been used as conditioning for camera pose estimation (e.g., Align3R~\cite{lu2025align3r}, Madpose~\cite{yu2025relative}).
In this spirit, we design a model that benefits from priors when available, yet remains well-posed without them.

\paragraph{Test-time Compute.}
Scaling test-time compute~\cite{wei2022chain, jaech2024openai, guo2025deepseek}
has emerged as an effective way to trade latency for accuracy in LLMs~\cite{achiam2023gpt, liu2024deepseek}.
Inspired by theories of fast and slow thinking~\cite{kahneman2011thinking}, a strong pre-trained model acts as fast "System‑1," while iterative reasoning at inference serves as "System‑2."
With techniques such as chain‑of‑thought and reflection, models revisit intermediate steps to refine predictions.
Beyond language, recent work brings this reasoning into continuous latent spaces~\cite{hao2024training}.
Another line of work views reasoning as an energy minimization process~\cite{du2022learning, du2024learning, gladstone2025energy}.
Energy‑based Transformers (EBTs)~\cite{gladstone2025energy}, for example, score the compatibility between inputs and predictions and use gradient‑guided sampling at test time to improve consistency.
Our approach shares this spirit: we define a test‑time objective that checks mutual consistency among outputs and refine all predictions to better satisfy it.
In vision, related efforts~\cite{gandelsman2022test} apply self‑supervised reconstruction losses at inference—e.g., masked‑autoencoder style objectives—to boost downstream performance.
More specifically in 3D vision,
Test3R~\cite{yuan2025test3r} fine-tunes DUSt3R~\cite{wang2023dust3r} at test time to generalize it from two-view modeling to multiple views.
TTT3R~\cite{chen2025ttt3r} proposes a test-time intervention mechanism to better generalize ~\cite{wang2025continuous} to long video sequences.
We follow the same spirit to frame 3D reconstruction/camera pose estimation with priors as a test-time constrained optimization problem.

%% file: sec/3_method.tex
\section{Methodology}
\label{sec:method}
\begin{figure*}[t]
    \centering
    \includegraphics[width=0.95\textwidth]{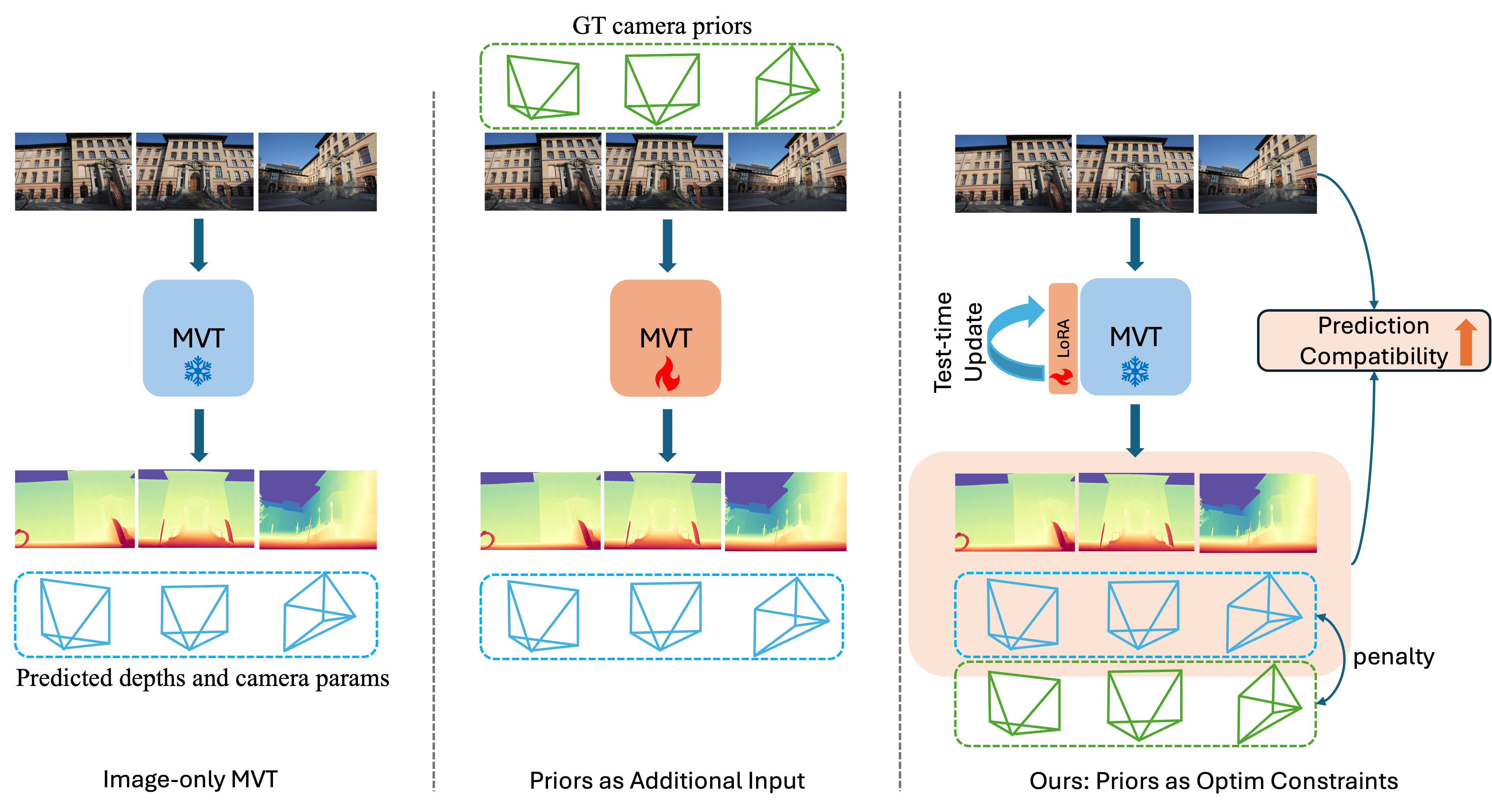}
    \caption{\textbf{Method Overview.} \textbf{(left)} Multi-view Transformers (MVTs) take a set of RGB images as input and output depth maps, camera poses, and intrinsics.
    \textbf{(middle)} Given camera priors, MapAnything~\cite{keetha2025mapanything} and Pow3R~\cite{jang2025pow3r} feed them into the network as additional input modalities, which requires retraining a modified MVT.
    \textbf{(right)} Our method, Test-time Constrained Optimization (TCO), treats the priors as constraints on the MVT's predictions and optimizes the network with LoRA using both prior penalty terms and a self-supervised objective, namely MVT prediction compatibility, at inference time.}
    \vspace{-0.5em}
    \label{fig:method}
\end{figure*}

\subsection{Preliminary: Feed-forward 3D Reconstruction}
\paragraph{No Priors.} We define the feed-forward 3D reconstruction without auxiliary priors as follows:
The input is a set of RGB images $\mathbf{I} = \{\mathbf{I}_i\}_{i=1}^N$,
representing the same scene from different viewpoints.
The goal is to learn a mapping from the input to a corresponding set of geometric predictions.
For each input view $\mathbf{I}_i$, the output predictions include
camera pose $\mathbf{T}_i$, intrinsic matrix $\mathbf{K}_i$,
and depth map $\mathbf{D}_i$ with an associated confidence map $\mathbf{C}_i$.
Although different works output different geometric modalities,
we simplify them to the above three predictions for convenience, without loss of generality.
This is because the other predicted geometric modalities, like point maps, can be derived from these three predictions.
Specifically, a camera-coordinate point map~\cite{wang2025pi} can be derived from the depth map and intrinsics,
while world-coordinate point maps~\cite{wang2025vggt,wang2025continuous}
can be derived by composing the depth maps, intrinsics, and camera poses altogether.
Feed-forward 3D reconstruction is usually implemented with a multi-view transformer~\cite{vaswani2017attention} (MVT).
Formally, we have:
\begin{equation}
   \{(\mathbf{D}_i, \mathbf{T}_i, \mathbf{K}_i, \mathbf{C}_i)\}_{i=1}^N = \mathtt{MVT}_\theta(\mathbf{I}), \label{eq:unconstrained_mvt}
\end{equation}
The depth map $\mathbf{D}_i$ and the confidence map $\mathbf{C}_i$ have the same spatial resolution $H \times W$ as the input image $\mathbf{I}_i$.
The parameterization of camera poses $\mathbf{T}_i$ varies across different works.
For example,
VGGT~\cite{wang2025vggt} parameterizes $\mathbf{T}_i$ as a quaternion $\mathbf{q}_i \in \mathbb{R}^4$ and a translation vector $\mathbf{t}_i \in \mathbb{R}^3$,
while $\pi^3$~\cite{wang2025pi} parameterizes $\mathbf{T}_i$ as a rotation matrix $\mathbf{R}_i \in \mathbb{R}^{3 \times 3}$ and a translation vector $\mathbf{t}_i$.
Following most of the literature, we assume a pinhole camera model with the principal point at the center of the image.
Thus, the intrinsic matrix $\mathbf{K}_i$ is fully determined by the focal lengths $(f_x, f_y)$, i.e., $\mathbf{K}_i = \text{diag}(f_x, f_y, 1)$.
\paragraph{With Priors.} In this case,
the input is no longer just a set of RGB images $\mathbf{I}$,
but may also include optional input modalities.
Both Pow3R~\cite{jang2025pow3r} and MapAnything~\cite{keetha2025mapanything} support input modalities including camera poses,
intrinsics, and depth maps.
To incorporate the auxiliary modalities,
they are first encoded by shallow ConvNets or MLPs $\mathtt{Enc}$,
and then added to the features in the forward pass of the MVT.
Formally, we have:
\begin{equation}
    \{(\mathbf{D}_i, \mathbf{T}_i, \mathbf{K}_i, \mathbf{C}_i)\}_{i=1}^N = \mathtt{MVT}_\theta(\mathbf{I}, \mathtt{Enc}(\mathbf{[D, T, K]})). \label{eq:constrained_mvt}
\end{equation}
The predictions are the same as in the unconstrained case.
\subsection{Test-time Constrained Optimization (TCO)}
Although both Pow3R and MapAnything achieve improved performance after fusing the auxiliary modalities with RGB images,
they modify the unconstrained MVT architecture to accept the auxiliary modalities,
hindering the reuse of pre-trained unconstrained MVTs, such as VGGT~\cite{wang2025vggt}.
As shown in Fig.~\ref{fig:method}, we offer a different perspective on incorporating the auxiliary modalities:
Instead of adding them as input to the MVT,
we formulate feed-forward 3D reconstruction with priors as a \textit{Test-time Constrained Optimization} (TCO) problem.
Specifically, each prior is treated as a constraint $g(\cdot)$ on the predictions,
for example, the camera pose prior can be expressed as $g=\|\mathbf{T}_i - \mathbf{T}_i^\text{prior}\| = 0$.
Given $M$ such priors, we formulate the TCO problem as:
\begin{equation}
\begin{alignedat}{2}
    & \min_{\theta} &\quad& \mathcal{L}(\mathtt{MVT}_\theta(\mathbf{I})) \\ \label{eq:const_opt}
    & \text{s.t.}   &\quad& g_m(\mathtt{MVT}_\theta(\mathbf{I})) = 0 \quad \forall m = 1, \ldots, M
\end{alignedat}
\end{equation}
where $\mathcal{L}(\cdot)$ is a self-supervised loss function that should strongly correlate with reconstruction quality.
We transform Eq.~\ref{eq:const_opt} into an unconstrained optimization problem with penalty terms~\cite{nocedal2006numerical}:
\begin{equation}
    \min_{\theta} \mathcal{L}(\mathtt{MVT}_\theta(\mathbf{I})) + \sum_{m=1}^M \mu_m \| g_m(\mathtt{MVT}_\theta(\mathbf{I})) \| \label{eq:unconstr_opt}
\end{equation}
where $\mu_m$ is a hyperparameter representing the penalty severity for violating the $m$-th constraint.
Since Eq.~\ref{eq:unconstr_opt} is non-linear and non-convex,
we solve it using gradient-based methods.
We will next introduce how to design the objective function $\mathcal{L}(\cdot)$ and the constraints $g_m(\cdot)$.
\subsubsection{Prediction Compatibility as Objective Function}
We use the MVTs' prediction compatibility as the objective function $\mathcal{L}$ to guide the optimization.
By prediction compatibility, we mean the consistency across the modalities of the MVTs' predictions.
For example, combining the depth maps from one scene with the camera poses from another scene will yield incorrect 3D reconstruction results.
Thus, the compatibility between predictions is a necessary condition for a high-quality 3D reconstruction.
We encourage prediction compatibility by minimizing the photometric and geometric errors between the renderings projected from other views and each view itself.
We first randomly split the input views into source and target groups $\mathbf{I}_{src}$ and $\mathbf{I}_{tgt}$.
The output predictions for the source views and target views are $\{\mathbf{D}_{src}, \mathbf{T}_{src}, \mathbf{K}_{src}, \mathbf{C}_{src}\}$
and $\{\mathbf{D}_{tgt}, \mathbf{T}_{tgt}, \mathbf{K}_{tgt}, \mathbf{C}_{tgt}\}$, respectively.
We use 2DGS~\cite{kerbl20233d, huang20242d} as a differentiable renderer to project the source RGB images and depth maps to the target views using the predicted source and target camera poses and intrinsics.
We choose 2DGS because it is designed to better model scene geometry.
Rather than optimizing 2DGS parameters directly,
we use a heuristic rule to infer the 2DGS parameters from the MVTs' predictions,
leaving the adaptation to the MVT parameters:
\begin{equation*}
\begin{aligned}
    c_{i,x,y} & = \mathbf{I}_{i}[x, y], \\
    \boldsymbol{\mu}_{i,x,y} & = \mathbf{T}_{i} \mathbf{p}_{i,x,y}, \\
    o_{i,x,y} & = 1 - 1/\mathbf{C}_{i,x,y}, \\
    \mathbf{n}_{i,x,y} & = \frac{\nabla_x \mathbf{p}_{i,x,y}\times\nabla_y\mathbf{p}_{i,x,y}}{\|\nabla_x \mathbf{p}_{i,x,y} \times \nabla_y\mathbf{p}_{i,x,y}\|}, \\
    \mathbf{q}_{i,x,y} & = \mathtt{quat}(\frac{\nabla_x \mathbf{p}_{i,x,y}}{\|\nabla_x \mathbf{p}_{i,x,y}\|}, \frac{\mathbf{n}_{i,x,y} \times \nabla_x \mathbf{p}_{i,x,y}}{\|\mathbf{n}_{i,x,y} \times \nabla_x \mathbf{p}_{i,x,y}\|}, \mathbf{n}_{i,x,y}), \\
    \mathbf{r}_{i,x,y} & = |\mathbf{n}_{i,x,y}[z]| \cdot [\|\nabla_x \mathbf{p}_{i,x,y}\|, \|\nabla_y \mathbf{p}_{i,x,y}\|] \\
\end{aligned}
\end{equation*}
where $\mathbf{I}_i \in \mathbf{I}_{src}$,
$\mathbf{p}_{i,x,y} = \mathbf{K}_{i}^{-1} [x, y, 1]^T \mathbf{D}_{i,x,y}$ is the 3D point in the source view $i$ at pixel $(x, y)$.
We use the predicted world-coordinate points as the centers $\boldsymbol{\mu}$ of the 2D Gaussians,
the pixel colors as the 2DGS colors $c$,
and the surface normals as the 2DGS normals $\mathbf{n}$, which are derived from the cross product of the point-map gradients.
Then, the 2DGS quaternion $\mathbf{q}$ is determined by the normalized x-axis point-map gradient, the corresponding normalized bitangent, and the surface normal vector $\mathbf{n}$.
The opacity $o$ is proportional to the predicted confidence.
When the predicted confidence is low,
the 2DGS is more transparent,
resulting in a lower contribution to the final radiance field.
Since $\exp(\cdot)+1$ is used as the confidence activation function,
we always have $\mathbf{C}_{i,x,y} > 1$.
Thus, we define the opacity as $1 - 1/\mathbf{C}_{i,x,y}$.
We define the radius $\mathbf{r}$ based on the magnitudes of the point-map gradients and further modulate it by the absolute z-component of the surface normal.
The gradient magnitudes reflect the local sampling density: larger gradients indicate that the region is more sparsely sampled, so we use a larger radius to cover it.
The normal term also accounts for view-dependent reliability: when the surface normal is nearly orthogonal to the viewing direction, the surface is observed at a grazing angle and its geometry is typically noisier than that of front-facing regions.
Therefore, the absolute z-component shrinks the projected support for such regions.

After obtaining the source views' 2DGS parameters,
we can render RGB images and depth maps in the target views using 2DGS rasterization~\cite{pfister2000surfels,huang20242d}.
We define the photometric and geometric losses as:
\begin{equation}
    \begin{aligned}
    \mathcal{L}_{\mathtt{photo}}(\mathbf{I}_{src}, \{\mathbf{I}_j\}_{j=1}^{N_{tgt}}) = \sum_{j=1}^{N_{tgt}} & \left\|\mathbf{I}_{j} - \mathbf{I}_{src\rightarrow j}\right\|_1 \\
    \mathcal{L}_{\mathtt{geom}}(\mathbf{D}_{src}, \{\mathbf{D}_j\}_{j=1}^{N_{tgt}}) = \sum_{j=1}^{N_{tgt}} & \left\|\mathbf{D}_{j} - \mathbf{D}_{src\rightarrow j}\right\|_1
    \end{aligned}
\end{equation}
where $\mathbf{I}_{src\rightarrow j}$ and $\mathbf{D}_{src\rightarrow j}$ are the RGB image and depth map rendered from the source views into target view $j$.
In practice,
we use the photometric loss as the SSL objective in point-map estimation and the geometric loss in camera pose estimation.
\subsubsection{Priors as Constraints}
We have transformed the constraints $g_m(\cdot)$ into penalty terms $\| g_m(\cdot) \|$ in Eq.~\ref{eq:unconstr_opt},
i.e., losses between the priors and the MVTs' corresponding predictions.
Now, we formulate the loss for each considered prior.
\\
\textbf{Camera Pose.} We first express all camera pose priors in the first-view coordinate system,
so that the first-view pose becomes $\hat{\mathbf{T}}_1 = [\mathbf{I} | \mathbf{0}]$,
aligning with the predicted camera pose format.
Given the scale ambiguity of the camera pose,
we express it as $[\mathbf{R}_i | \mathbf{t}_i]$ and decompose the loss into rotation and translation terms.
We use the angular distance to measure the rotation difference as follows:
\begin{equation}
    g_{\mathtt{rot}}(\mathbf{R}_i; \hat{\mathbf{R}}_i) = \arccos\big(\frac{\text{trace}(\mathbf{R}_i^T \hat{\mathbf{R}}_i) - 1}{2}\big)
\end{equation}
For the translation term,
we estimate the scene scale as the mean distance from camera centers to the reference origin,
computed separately for the predicted poses ($s$) and the pose priors ($\hat{s}$).
We then normalize the translation vectors by their respective scales and define the loss
as the $\ell_1$ distance between the normalized translations:
\begin{equation}
    g_{\mathtt{trans}}(\mathbf{t}_i; \hat{\mathbf{t}}_i) = \left\| \frac{\mathbf{t}_i}{s} - \frac{\hat{\mathbf{t}}_i}{\hat{s}} \right\|_1
\end{equation}
\\
\textbf{Intrinsics.} Under a pinhole camera model with the principal point at the image center,
we penalize discrepancies between the predicted and prior focal lengths.
\begin{equation}
    g_{\mathtt{K}}(\mathbf{K}_i; \hat{\mathbf{K}}_i) = \left\| f_x - \hat{f}_x \right\|_1 + \left\| f_y - \hat{f}_y \right\|_1
\end{equation}
where $(f_x, f_y)$ and $(\hat{f}_x, \hat{f}_y)$ are given by the diagonal entries of the intrinsic matrices $\mathbf{K}_i$ and $\hat{\mathbf{K}}_i$, respectively.
\\
\textbf{Depth Map.}
We first compute a global scale $\mathtt{s}$ and shift $\mathtt{t}$ to align the predicted depth maps $\mathbf{D} = \{\mathbf{D}_i\}_{i=1}^N$ with the priors $\hat{\mathbf{D}} = \{\hat{\mathbf{D}}_i\}_{i=1}^N$ by solving a least-squares problem over all views in the same scene:
\begin{equation}
\begin{aligned}
    \mathtt{s} &= \frac{\sum_{i,x,y} (\mathbf{D}_i[x, y] - \bar{\mathbf{D}})(\hat{\mathbf{D}}_i[x, y] - \bar{\hat{\mathbf{D}}})}
    {\sum_{i,x,y} (\mathbf{D}_i[x, y] - \bar{\mathbf{D}})^2} \\
    \mathtt{t} &= \bar{\hat{\mathbf{D}}} - \mathtt{s} \cdot \bar{\mathbf{D}}
\end{aligned}
\end{equation}
where $\bar{\mathbf{D}}$ and $\bar{\hat{\mathbf{D}}}$ denote the averages over all pixels from all views.
After estimating the global alignment parameters $\mathtt{s}$ and $\mathtt{t}$, we apply them to each view individually and define the per-view depth loss as:
\begin{equation}
    g_{\mathtt{depth}}(\mathbf{D}_i; \hat{\mathbf{D}}_i) = \left\| \mathtt{s} \cdot \mathbf{D}_i + \mathtt{t} - \hat{\mathbf{D}}_i \right\|_1
\end{equation}

\subsubsection{Test-time Training Loss}
After defining the objective function and the constraints,
we next combine them to form the test-time training loss.
Empirically, we find that the photometric loss is most effective for point-map estimation,
whereas the geometric loss is most effective for camera pose estimation.
Accordingly, we define the test-time training loss for each task as follows.
\\
\textbf{Task 1: Point Map Estimation with Priors.}
In this task, we assume the camera poses and/or intrinsics are known.
Thus, the test-time training loss is:
\begin{equation}
    \begin{aligned}
    \mathcal{L}_{\mathtt{TCO}} = &\lambda_1 \mathcal{L}_{\mathtt{photo}} + \mu_1 \sum_{i=1}^N g_{\mathtt{rot}}(\mathbf{R}_i; \hat{\mathbf{R}}_i) \\
    &+ \mu_2 \sum_{i=1}^N g_{\mathtt{trans}}(\mathbf{t}_i; \hat{\mathbf{t}}_i) + \mu_3 \sum_{i=1}^N g_{\mathtt{K}}(\mathbf{K}_i; \hat{\mathbf{K}}_i)
    \end{aligned}
\end{equation}
where $\lambda_1$ is the weight for the photometric loss,
$\mu_1$, $\mu_2$, and $\mu_3$ are the weights for the rotation, translation, and intrinsic losses.
\\
\textbf{Task 2: Camera Pose Estimation with Priors.}
In this task, we assume the depth maps are known.
Thus, the test-time training loss is:
\begin{equation}
    \begin{aligned}
    \mathcal{L}_{\mathtt{TCO}} = &\lambda_1 \mathcal{L}_{\mathtt{geom}} + \mu_1 \sum_{i=1}^N g_{\mathtt{depth}}(\mathbf{D}_i; \hat{\mathbf{D}}_i)
    \end{aligned}
\end{equation}
where $\lambda_1$ is the weight for the geometric loss,
$\mu_1$ is the weight for the depth loss.

\subsection{Fine-tuning Strategy for Task Synergy}
To better exploit the synergy across different prediction tasks, we adopt a selective fine-tuning strategy.
Our goal is not merely to fit the priors, but to let the prior-constrained tasks improve the shared scene representation in a way that also benefits the unconstrained predictions.
To this end, we freeze all task-specific heads to avoid overfitting to individual priors, while fine-tuning only the shared decoder network.
Ablation studies show that this fine-tuning strategy is crucial to the success of our method.

\subsection{Implementation Details}
We use Adam~\cite{kingma2014adam} as the test-time optimizer.
We use LoRA~\cite{hu2022lora} to fine-tune the shared decoder network in a parameter-efficient manner.
LoRA is applied to all qkv (query, key, value) projections and feed-forward layers in the decoder.
We set the rank of the LoRA matrices to 4.
The number of test-time iterations usually ranges from 5 to 80.
To improve test-time training efficiency,
we freeze the entire encoder network.
Following the strategy described above, we also freeze all task-specific heads.
We use gsplat~\cite{ye2025gsplat} as the differentiable renderer.
In practice, for each test-time iteration, we randomly sample one view as the source and render it into all other views.
Please refer to the Supplementary Material for additional implementation details.

%% file: sec/4_experiments.tex
\section{Experiments}
We evaluate our method on two tasks:
point-map estimation and camera pose estimation, respectively with camera parameters and depth maps as priors.
Across all benchmarks, we observe consistent improvements over different base MVTs,
i.e., VGGT~\cite{wang2025vggt} and $\pi^3$~\cite{wang2025pi}.
\subsection{Point Map Estimation with Priors}
\subsubsection{Evaluation Settings}
We follow the same evaluation settings as $\pi^3$~\cite{wang2025pi},
which evaluates the quality of reconstructed point maps on four datasets: ETH3D~\cite{schops2017multi} (outdoor scenes),
DTU~\cite{jensen2014large} (object-centric),
7-Scenes~\cite{shotton2013scene} (indoor scenes),
and NRGBD~\cite{azinovic2022neural} (indoor scene).
On the ETH3D and DTU datasets,
keyframes are sampled with a stride of 5 images.
On 7-Scenes and NRGBD datasets,
we only adopt the sparse view setting due to memory constraints,
which samples keyframes with a stride of 200 (7-Scenes) or 500 (NRGBD).
We align the post-processing of reconstructed point maps with $\pi^3$~\cite{wang2025pi},
which uses the Umeyama algorithm for a coarse Sim(3) alignment,
followed by refinement with the Iterative Closest Point (ICP) algorithm.
Following prior works,
we report Accuracy (Acc.), Completion (Comp.), and Normal Consistency (N.C.) metrics.
\subsubsection{Performance.}
\begin{table}[t]
    \centering
    \caption{
        \textbf{Point Map Estimation on ETH3D.}
        TCO outperforms both base image-only MVTs and prior-aware feed-forward methods, achieving the best overall performance.
        Camera poses are the most important prior for improving 3D reconstruction on ETH3D, and combining them with intrinsics yields further gains.
    }
    \vspace{-0.5em}
    \tablestyle{2pt}{1.0}
    \resizebox{1.0\columnwidth}{!}{
    \begin{tabular}{lccc@{\hspace{0.8em}}cccccc}
        \toprule[0.1em]
        {\multirow{2}{*}{\textbf{Method}}} &
        {\multirow{2}{*}{\textit{Compat}}} &
        {\multirow{2}{*}{\textit{Pose}}} &
        {\multirow{2}{*}{\textit{Intr}}} &
        \multicolumn{2}{c}{\textbf{Acc.} $\downarrow$}  &
        \multicolumn{2}{c}{\textbf{Comp.} $\downarrow$} &
        \multicolumn{2}{c}{\textbf{N.C.} $\uparrow$} \\
        \cmidrule(r){5-6} \cmidrule(r){7-8} \cmidrule(r){9-10}
        & & & & Mean & Med. & Mean & Med. & Mean & Med. \\
        \midrule[0.08em]
        Fast3R~\cite{yang2025fast3r} & \xmark & \xmark & \xmark & 0.832 & 0.691 & 0.978 & 0.683 & 0.667 & 0.766 \\
        CUT3R~\cite{wang2025continuous} & \xmark & \xmark & \xmark & 0.617 & 0.525 & 0.747 & 0.579 & 0.754 & 0.848 \\
        FLARE~\cite{zhang2025flare} & \xmark & \xmark & \xmark & 0.464 & 0.338 & 0.664 & 0.395 & 0.744 & 0.864 \\
        \midrule[0.08em]
        Pow3R (\textit{asym.})~\cite{jang2025pow3r} & \xmark & \checkmark & \checkmark & 0.230 & 0.146 & 0.188 & 0.119 & 0.865 & 0.968 \\
        MapAnything~\cite{keetha2025mapanything} & \xmark & \checkmark & \checkmark & 0.167 & 0.100 & 0.131 & 0.066 & 0.866 & 0.968 \\
        \midrule[0.08em]
        VGGT~\cite{wang2025vggt} & \xmark & \xmark & \xmark & 0.280 & 0.185 & 0.305 & 0.182 & 0.853 & 0.950 \\
        VGGT raw depth & \xmark & \checkmark & \checkmark & 0.283 & 0.191 & 0.330 & 0.203 & 0.834 & 0.931 \\
        TCO-VGGT & \checkmark & \xmark & \xmark & 0.270 & 0.178 & 0.298 & 0.179 & 0.853 & 0.949 \\
        TCO-VGGT & \checkmark & \xmark & \checkmark & 0.273 & 0.175 & 0.302 & 0.181 & 0.839 & 0.938 \\
        TCO-VGGT & \checkmark & \checkmark & \xmark & 0.185 & 0.121 & 0.179 & 0.103 & 0.875 & 0.972 \\
        TCO-VGGT & \xmark & \checkmark & \checkmark & 0.134 & 0.079 & 0.149 & 0.083 & 0.884 & 0.977 \\    
        TCO-VGGT & \checkmark & \checkmark & \checkmark & \underline{0.114} & \underline{0.060} & \underline{0.116} & \underline{0.054} & \underline{0.901} & \underline{0.984} \\
        \midrule[0.08em]
        $\pi^3$~\cite{wang2025pi} & \xmark & \xmark & \xmark & 0.194 & 0.131 & 0.210 & 0.128 & 0.883 & 0.969 \\
        TCO-$\pi^3$ & \checkmark & \checkmark & \checkmark & \textbf{0.071} & \textbf{0.040} & \textbf{0.065} & \textbf{0.037} & \textbf{0.916} & \textbf{0.985} \\
        \bottomrule[0.1em]
    \end{tabular}
    }
    \vspace{-0.65em}
    \label{tab:mv_recon_eth3d}
\end{table}
\begin{table}[ht]
    \centering
    \caption{
        \textbf{Point Map Estimation on DTU.} TCO outperforms both the base image-only models and the prior-aware feed-forward methods.
        The proposed prediction compatibility loss is essential for incorporating priors as a regularization signal from distillation priors into the network.
    }
    \vspace{-0.5em}
    \tablestyle{2pt}{1}
    \resizebox{1.0\columnwidth}{!}{
    \begin{tabular}{lccc@{\hspace{0.8em}}cccccc}
        \toprule[0.1em]
        {\multirow{2}{*}{\textbf{Method}}} &
        {\multirow{2}{*}{\textit{Compat}}} &
        {\multirow{2}{*}{\textit{Pose}}} &
        {\multirow{2}{*}{\textit{Intr}}} &
        \multicolumn{2}{c}{\textbf{Acc.} $\downarrow$}  &
        \multicolumn{2}{c}{\textbf{Comp.} $\downarrow$} &
        \multicolumn{2}{c}{\textbf{N.C.} $\uparrow$} \\
        \cmidrule(r){5-6} \cmidrule(r){7-8} \cmidrule(r){9-10}
        & & & & Mean & Med. & Mean & Med. & Mean & Med. \\
        \midrule[0.08em]
        Fast3R~\cite{yang2025fast3r} & \xmark & \xmark & \xmark & 3.340 & 1.919 & 2.929 & 1.125 & 0.671 & 0.755 \\
        CUT3R~\cite{wang2025continuous} & \xmark & \xmark & \xmark & 4.742 & 2.600 & 3.400 & 1.316 & 0.679 & 0.764 \\
        FLARE~\cite{zhang2025flare} & \xmark & \xmark & \xmark & 2.541 & 1.468 & 3.174 & 1.420 & \textbf{0.684} & \textbf{0.774} \\
        \midrule[0.08em]
        Pow3R~\cite{jang2025pow3r} & \xmark & \checkmark & \checkmark & 1.234 & 0.620 & 2.357 & 0.833 & 0.680 & 0.765 \\
        MapAnything~\cite{keetha2025mapanything} & \xmark & \checkmark & \checkmark & 2.731 & 1.570 & 2.209 & 0.892 & \underline{0.681} & 0.767 \\
        \midrule[0.08em]
        VGGT~\cite{wang2025vggt} & \xmark & \xmark & \xmark & 1.338 & 0.779 & 1.896 & 0.992 & 0.676 & \textbf{0.766} \\
        TCO-VGGT & \checkmark & \xmark & \xmark & 1.170 & 0.670 & 1.591 & 0.734 & 0.673 & 0.761 \\
        TCO-VGGT & \xmark & \checkmark & \checkmark & 1.327 & 0.747 & 1.968 & 1.022 & 0.669 & 0.754 \\
        TCO-VGGT & \checkmark & \checkmark & \checkmark & \textbf{0.976} & \textbf{0.542} & \textbf{1.299} & \textbf{0.598} & \textbf{0.677} & 0.765 \\
        \midrule[0.08em]
        $\pi^3$~\cite{wang2025pi} & \xmark & \xmark & \xmark & 1.198 & 0.646 & 1.849 & \underline{0.607} & 0.678 & \underline{0.768} \\
        TCO-$\pi^3$ & \checkmark & \checkmark & \checkmark & \underline{1.109} & \underline{0.595} & \underline{1.524} & 0.646 & 0.670 & 0.756 \\
        \bottomrule[0.1em]
    \end{tabular}
    }
    \vspace{-0.65em}
    \label{tab:mv_recon_dtu}
\end{table}

\paragraph{Comparison with Image-only Models.}
In general,
we observe consistent improvements over the base image-only models across all benchmarks.
In Tab.~\ref{tab:mv_recon_eth3d}, we report the details of how much each prior contributes to the improvement on ETH3D.
We find that camera poses are the most important prior for improving 3D reconstruction quality on ETH3D.
Compared with camera poses, intrinsics alone, even with the self-supervised objective, do not help.
However, when combined with camera poses, intrinsics can further improve the reconstruction quality.
Even without the compatibility objective, camera poses plus intrinsics can already improve the reconstruction quality by a large margin.
The \textit{VGGT raw depth} entry is a baseline that directly combines the raw depth predicted by VGGT with prior camera parameters after scale alignment.
It shows that the gains of TCO come from geometry refinement rather than simply re-stitching raw depth maps.
On the DTU dataset (Tab.~\ref{tab:mv_recon_dtu}), the trend is different.
Without the compatibility objective, camera poses plus intrinsics deteriorate the reconstruction performance.
However, when combined with the compatibility objective, camera poses plus intrinsics can improve the reconstruction quality by a large margin,
proving the necessity and effectiveness of the compatibility objective for incorporating priors as a regularization of distillation priors into the network.
Interestingly, we find that the SSL objective alone can already improve the reconstruction quality without any priors on the DTU dataset,
which suggests a possible post-processing approach for self-improving 3D reconstruction.
Similar phenomena are also observed on the 7-Scenes (Tab.~\ref{tab:mv_recon_7scenes}) and NRGBD (Tab.~\ref{tab:mv_recon_nrgbd}) datasets,
where TCO can consistently boost VGGT's performance by incorporating camera priors.
To demonstrate the architecture-agnostic nature of TCO,
we also provide results with another base MVT model, $\pi^3$~\cite{wang2025pi}.
Again, we observe consistent and significant improvements across all benchmarks.
In addition,
we report results for other recent image-only models (Fast3R~\cite{yang2025fast3r}, CUT3R~\cite{wang2025continuous}, FLARE~\cite{zhang2025flare}) for a comprehensive comparison.
\begin{table}[t]
    \centering
    \caption{
        \textbf{Point Map Estimation on 7-Scenes (sparse view).} Our method outperforms both the base image-only models and the prior-aware feed-forward methods.
    }
    \vspace{-0.5em}
    \tablestyle{2pt}{1.0}
    \resizebox{1.0\columnwidth}{!}{
    \begin{tabular}{lccc@{\hspace{0.8em}}cccccc}
        \toprule[0.1em]
        {\multirow{2}{*}{\textbf{Method}}} &
        {\multirow{2}{*}{\textit{Compat}}} &
        {\multirow{2}{*}{\textit{Pose}}} &
        {\multirow{2}{*}{\textit{Intr}}} &
        \multicolumn{2}{c}{\textbf{Acc.} $\downarrow$}  &
        \multicolumn{2}{c}{\textbf{Comp.} $\downarrow$} &
        \multicolumn{2}{c}{\textbf{NC.} $\uparrow$} \\
        \cmidrule(r){5-6} \cmidrule(r){7-8} \cmidrule(r){9-10}
         &  &  &  & Mean & Med. & Mean & Med. & Mean & Med. \\
        \midrule[0.08em]
        Fast3R~\cite{yang2025fast3r} & \xmark & \xmark & \xmark & 0.096 & 0.065 & 0.145 & 0.093 & 0.672 & 0.760 \\
        CUT3R~\cite{wang2025continuous} & \xmark & \xmark & \xmark & 0.094 & 0.051 & 0.101 & 0.050 & 0.703 & 0.804 \\
        FLARE~\cite{zhang2025flare} & \xmark & \xmark & \xmark & 0.085 & 0.058 & 0.142 & 0.104 & 0.695 & 0.779 \\
        \midrule[0.08em]
        Pow3R~\cite{jang2025pow3r} & \xmark & \checkmark & \checkmark & 0.034 & 0.019 & 0.041 & 0.022 & 0.763 & 0.880 \\
        MapAnything~\cite{keetha2025mapanything} & \xmark & \checkmark & \checkmark & 0.026 & 0.015 & 0.032 & 0.018 & 0.773 & 0.889 \\
        \midrule[0.08em]
        VGGT~\cite{wang2025vggt} & \xmark & \xmark & \xmark & 0.046 & 0.026 & 0.057 & 0.034 & 0.728 & 0.842 \\
        TCO-VGGT & \checkmark & \checkmark & \checkmark & \textbf{0.020} & \textbf{0.012} & \underline{0.026} & \textbf{0.015} & \underline{0.790} & \textbf{0.908} \\
        \midrule[0.08em]
        $\pi^3$~\cite{wang2025pi} & \xmark & \xmark & \xmark & 0.048 & 0.028 & 0.072 & 0.047 & 0.742 & 0.842 \\
        TCO-$\pi^3$ & \checkmark & \checkmark & \checkmark & \underline{0.020} & \underline{0.012} & \textbf{0.024} & \underline{0.015} & \textbf{0.795} & \underline{0.907} \\
        \bottomrule[0.1em]
    \end{tabular}
    }
    \vspace{-0.65em}
    \label{tab:mv_recon_7scenes}
\end{table}

\begin{table}[t]
    \centering
    \caption{
        \textbf{Point Map Estimation on NRGBD (sparse view).} Our method outperforms both the base image-only models and the prior-aware feed-forward methods.
    }
    \vspace{-0.5em}
    \tablestyle{2pt}{1.0}
    \resizebox{1.0\columnwidth}{!}{
    \begin{tabular}{lccc@{\hspace{0.8em}}cccccc}
        \toprule[0.1em]
        {\multirow{2}{*}{\textbf{Method}}} &
        {\multirow{2}{*}{\textit{Compat}}} &
        {\multirow{2}{*}{\textit{Pose}}} &
        {\multirow{2}{*}{\textit{Intr}}} &
        \multicolumn{2}{c}{\textbf{Acc.} $\downarrow$}  &
        \multicolumn{2}{c}{\textbf{Comp.} $\downarrow$} &
        \multicolumn{2}{c}{\textbf{NC.} $\uparrow$} \\
        \cmidrule(r){5-6} \cmidrule(r){7-8} \cmidrule(r){9-10}
         &  &  &  & Mean & Med. & Mean & Med. & Mean & Med. \\
        \midrule[0.08em]
        Fast3R~\cite{yang2025fast3r} & \xmark & \xmark & \xmark & 0.135 & 0.091 & 0.163 & 0.104 & 0.759 & 0.877 \\
        CUT3R~\cite{wang2025continuous} & \xmark & \xmark & \xmark & 0.104 & 0.041 & 0.079 & 0.031 & 0.822 & 0.968 \\
        FLARE~\cite{zhang2025flare} & \xmark & \xmark & \xmark & 0.053 & 0.024 & 0.051 & 0.025 & 0.877 & 0.988 \\
        \midrule[0.08em]
        Pow3R~\cite{jang2025pow3r} & \xmark & \checkmark & \checkmark & 0.062 & 0.029 & 0.062 & 0.029 & 0.844 & 0.983 \\
        MapAnything~\cite{keetha2025mapanything} & \xmark & \checkmark & \checkmark & 0.040 & 0.024 & 0.041 & 0.020 & 0.875 & 0.985 \\
        \midrule[0.08em]
        VGGT~\cite{wang2025vggt} & \xmark & \xmark & \xmark & 0.051 & 0.029 & 0.066 & 0.038 & 0.890 & 0.981 \\
        TCO-VGGT & \checkmark & \checkmark & \checkmark & \underline{0.023} & \underline{0.013} & \underline{0.027} & \underline{0.013} & \textbf{0.921} & \textbf{0.995} \\
        \midrule[0.08em]
        $\pi^3$~\cite{wang2025pi} & \xmark & \xmark & \xmark & 0.026 & 0.015 & 0.028 & 0.014 & \underline{0.916} & 0.992 \\
        TCO-$\pi^3$ & \checkmark & \checkmark & \checkmark & \textbf{0.020} & \textbf{0.011} & \textbf{0.022} & \textbf{0.010} & 0.915 & \underline{0.993} \\
        \bottomrule[0.1em]
    \end{tabular}
    }
    \vspace{-0.65em}
    \label{tab:mv_recon_nrgbd}
\end{table}

\paragraph{Comparison with Other Prior-aware Methods.}
Beyond comparing with the image-only models,
we also compare our method with recent prior-aware feed-forward methods, Pow3R~\cite{jang2025pow3r} and MapAnything~\cite{keetha2025mapanything}.
Both methods retrain a new model by feeding priors into the MVT with a modified architecture.
Surprisingly,
we find that our method achieves superior performance to these two methods across all benchmarks.
These results demonstrate the effectiveness of our test-time method in incorporating priors for 3D reconstruction.

\begin{figure*}[t]
    \centering
    \includegraphics[width=1.0\textwidth]{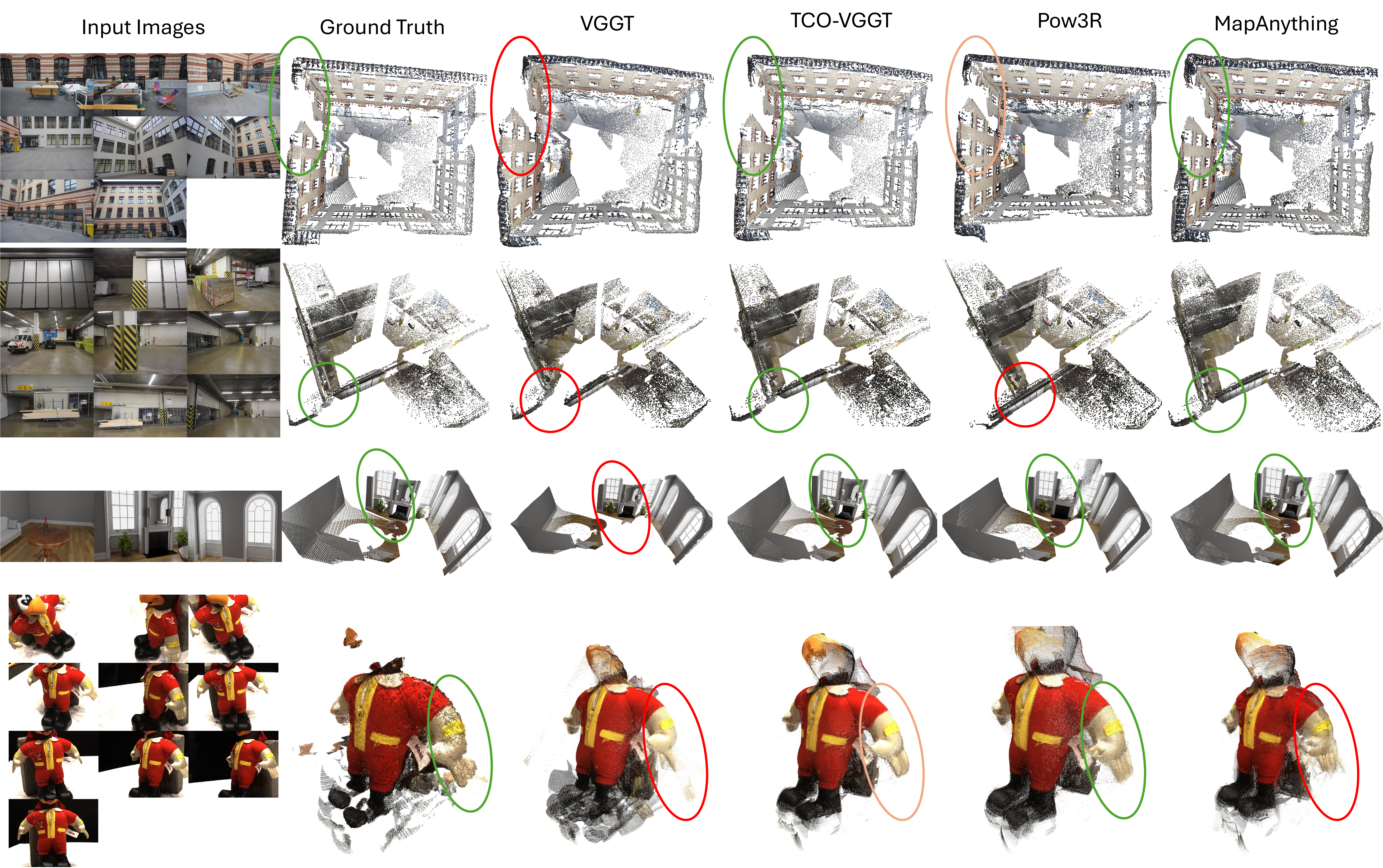}
    \caption{
        \textbf{Qualitative Comparison.}
        We compare TCO-VGGT with the base image-only model VGGT and the prior-aware feed-forward methods Pow3R and MapAnything.
        Overall, TCO-VGGT effectively corrects structural errors in image-only reconstructions by incorporating camera priors.
        \textcolor{red}{Red}, \textcolor{orange}{orange}, and \textcolor{green!60!black}{green} circles highlight regions that are wrongly reconstructed, partially corrected, and correctly reconstructed, respectively.
        In the first row, TCO-VGGT corrects the inaccurate relative positions of two walls in the VGGT reconstruction.
        The same phenomenon is observed in the second and third rows, where the inaccurate scene structures are corrected.
        In the last row, TCO-VGGT reconstructs the hand as a whole, whereas VGGT reconstructs it as separate parts.
        MapAnything reconstructs the hand with blurry boundaries.
        Additional fine-grained qualitative comparisons with MapAnything and Pow3R are provided in the Supplementary Material.}
    \vspace{-0.55em}
    \label{fig:qualitative_res}
\end{figure*}

\paragraph{Qualitative Results.}
In Fig.~\ref{fig:qualitative_res}, we show the qualitative results of TCO on four scenes from the ETH3D, NRGBD, and DTU datasets.
In all four scenes,
TCO corrects structural errors in the image-only reconstructions by incorporating priors.
In the first row, TCO corrects the inaccurate relative positions of two walls in the VGGT reconstruction.
The same thing happens in the second and third rows,
where the inaccurate scene structures are corrected.
In the last row, TCO reconstructs the hand as a whole, whereas the base model VGGT reconstructs it as separate parts.
All these results demonstrate the effectiveness of TCO in incorporating priors for 3D reconstruction.
We provide more detailed qualitative results in the Supplementary Material to compare TCO with MapAnything and Pow3R.

\subsection{Camera Pose Estimation with Priors}
\begin{table}[t]
    \centering
    \caption{
        \textbf{Camera Pose Estimation on ScanNet with different keyframe strides.}
        Our method outperforms both the base image-only models and the prior-aware feed-forward methods.
    }
    \vspace{-0.5em}
    \tablestyle{2pt}{1.0}
    \resizebox{1.0\columnwidth}{!}{
    \begin{tabular}{c l c c c c c}
        \toprule[0.1em]
        \textbf{Stride} & \textbf{Method} & \textit{Compat} & \textit{Depth} & \textbf{ATE} $\downarrow$ & \textbf{RPE trans} $\downarrow$ & \textbf{RPE rot} $\downarrow$ \\
        \midrule[0.08em]
        
          & Pow3R~\cite{jang2025pow3r} & \xmark & \checkmark & 0.0172 & 0.0258 & 0.4614 \\
        1   & VGGT~\cite{wang2025vggt} & \xmark & \xmark & 0.0094 & 0.0160 & 0.3781 \\
           & TCO-VGGT                & \checkmark & \checkmark & \textbf{0.0072} & \textbf{0.0139} & \textbf{0.3731} \\
        \midrule[0.08em]
          & Pow3R~\cite{jang2025pow3r} & \xmark & \checkmark & 0.0201 & 0.0335 & 0.7114 \\
        4   & VGGT~\cite{wang2025vggt} & \xmark & \xmark & 0.0127 & 0.0215 & \textbf{0.5706} \\
           & TCO-VGGT                & \checkmark & \checkmark & \textbf{0.0106} & \textbf{0.0190} & 0.6026 \\
        \midrule[0.08em]
         & Pow3R~\cite{jang2025pow3r} & \xmark & \checkmark & 0.0620 & 0.0965 & 1.6233 \\
        16   & VGGT~\cite{wang2025vggt} & \xmark & \xmark & 0.0357 & 0.0563 & 1.3749 \\
           & TCO-VGGT                & \checkmark & \checkmark & \textbf{0.0301} & \textbf{0.0468} & \textbf{1.3677} \\
        \bottomrule[0.1em]
    \end{tabular}
    }
    \vspace{-0.65em}
    \label{tab:pose_scannet_strides}
\end{table}
We further evaluate the camera pose estimation performance on ScanNet.
Following the evaluation settings of $\pi^3$~\cite{wang2025pi,wang2025continuous,zhang2024monst3r},
we report distance error metrics, including Absolute Trajectory Error (ATE), Relative Pose Error for translation (RPE trans), and Relative Pose Error for rotation (RPE rot).
Predicted camera trajectories are aligned with the ground truth via a Sim(3) transformation before calculating the errors.
Due to memory constraints, we cannot evaluate depth-prompted camera pose estimation with 90 input views as in~\cite{wang2025pi,zhang2024monst3r}.
Instead, we evaluate in a sparse-view setting with six input views.
We further report results of TCO with larger keyframe sampling strides, 4 and 16,
to investigate how well our method can generalize to wider camera baselines, which is a more challenging task.
In Tab.~\ref{tab:pose_scannet_strides},
we show that TCO can consistently improve the performance of VGGT across all keyframe strides.

\subsection{Ablation Studies}
We conduct ablation studies to evaluate the contribution of each component of TCO
and to explore its limits under longer test-time iterations.
More experiments can be found in the Supplementary Material.
\begin{table}[t]
    \centering
    \caption{
        \textbf{Ablations on ETH3D.} We study the fine-tuning strategy and the 2DGS heuristics used in the compatibility objective.}
    \vspace{-0.45em}
    \tablestyle{2pt}{1.0}
    \resizebox{1.0\columnwidth}{!}{
    \begin{tabular}{l@{\hspace{0.8em}}cc@{\hspace{0.8em}}cc@{\hspace{0.8em}}cc}
        \toprule[0.1em]
        {\textbf{Setting}} &
        \multicolumn{2}{c}{\textbf{Acc.} $\downarrow$}  &
        \multicolumn{2}{c}{\textbf{Comp.} $\downarrow$} &
        \multicolumn{2}{c}{\textbf{N.C.} $\uparrow$} \\
        \cmidrule(r){2-3} \cmidrule(r){4-5} \cmidrule(r){6-7}
         & Mean & Med. & Mean & Med. & Mean & Med. \\
        \midrule[0.08em]
        \multicolumn{7}{l}{\textit{Fine-tuning strategy}} \\
        Finetune decoder (Ours) & \textbf{0.114} & \textbf{0.060} & 0.116 & \textbf{0.054} & \textbf{0.901} & \textbf{0.984}\\
        + Camera head           & 0.158 & 0.107 & 0.191 & 0.113 & 0.864 & 0.966 \\
        + Depth head            & 0.162 & 0.081 & \textbf{0.098} & 0.054 & 0.850 & 0.954 \\
        + Camera + depth head & 0.183 & 0.088 & 0.195 & 0.116 & 0.817 & 0.937 \\
        \midrule[0.08em]
        \multicolumn{7}{l}{\textit{2DGS Heuristics}} \\
        TCO-VGGT (Ours) & \textbf{0.114} & \textbf{0.060} & 0.116 & \textbf{0.054} & \textbf{0.901} & \textbf{0.984} \\
        No Confidence as Opacity & 0.120 & 0.069 & 0.135 & 0.076 & 0.893 & 0.981 \\
        Learnable quats \& radii & 0.121 & 0.068 & \textbf{0.113} & 0.060 & 0.891 & 0.980 \\
        Direct 2DGS optim (VGGT frozen) & 0.895 & 0.771 & 0.809 & 0.476 & 0.565 & 0.598 \\
        \bottomrule[0.1em]
    \end{tabular}
    }
    \vspace{-0.6em}
    \label{tab:ablation_finetune_opacity}
\end{table}
\subsubsection{Test-time Scaling Curve}
We plot the test-time scaling curve of our TCO-VGGT model on ETH3D.
The performance improves as the number of test-time iterations increases.
On the one hand, with only 5 test-time iterations,
TCO already improves performance over the base image-only model.
On the other hand, with 80 test-time iterations,
TCO achieves the best performance among all tested settings,
although the improvement margin compared to 40 iterations is not that large.
\begin{figure}[ht]
    \centering
    \includegraphics[width=1.0\columnwidth, trim=7 9 7 7, clip]{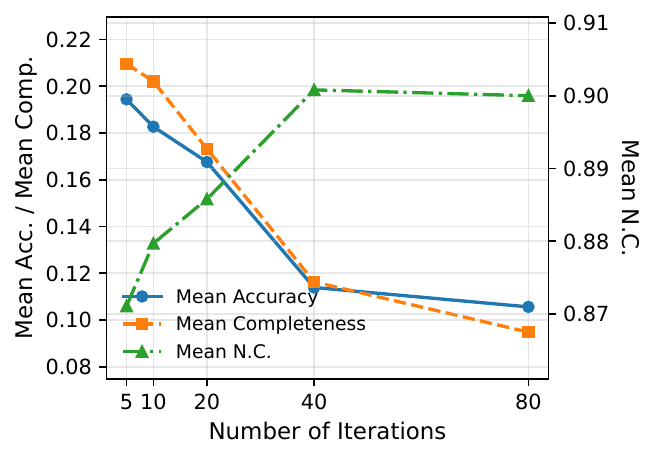}
    \caption{
        \textbf{Test-time scaling curve of our method on ETH3D.}
    }
    \vspace{-0.45em}
    \label{fig:test_time_scaling}
\end{figure}

\subsubsection{Fine-tuning Strategy}
We freeze all the prediction heads and only fine-tune the decoder network,
aiming to maximize the synergy between the different tasks by only adapting the shared decoder network.
To investigate the effectiveness of this fine-tuning strategy,
we compare the performance of TCO with and without fine-tuning the task heads.
We find that freezing the task heads is crucial to the success of TCO;
either fine-tuning the camera pose head or the depth head deteriorates the performance significantly.
This observation supports our design choice to maximize the synergy between tasks through shared-decoder adaptation.

\subsubsection{2DGS Heuristics}
We further ablate the 2DGS heuristics used in the cross-view compatibility objective.
As shown in Tab.~\ref{tab:ablation_finetune_opacity}, using confidence as opacity leads to slight improvements over the variant without this design.
By contrast, making the quaternions and radii learnable does not improve performance, while introducing additional optimization overhead.
Directly optimizing the 2DGS parameters while freezing VGGT performs substantially worse, 
indicating that adapting the MVT through our heuristic 2DGS parameterization is much more effective than optimizing the renderer itself.
We also study the role of z-component modulation in the radius, 
which yields similar results on ETH3D, NRGBD, and 7-Scenes, while achieving better results on DTU.

%% file: sec/5_conclusion.tex
\section{Conclusion}
We presented a test-time constrained optimization framework for multiview Transformers that incorporates camera and scene priors without retraining or architectural changes. 
By casting priors as differentiable penalty terms and optimizing a cross-view compatibility objective via 2DGS-based projections, 
our method exploits the synergy among depth, pose, and intrinsics while fine-tuning only the shared decoder. 
Extensive experiments on ETH3D, DTU, 7-Scenes, NRGBD, and ScanNet show consistent gains over base MVTs and 
even surpass prior-aware feed-forward baselines, with robustness under sparse views and wider baselines. 
This plug-and-play approach provides a practical path to leverage available priors at inference.

%% file: sec/6_ack.tex
\section{Acknowledgements}
This work was supported in part by the National Science Foundation (NSF) award IIS-2212046.

%% file: sec/X_suppl.tex
\clearpage
\setcounter{page}{1}
\maketitlesupplementary
\section{More Ablation Studies}
\subsection{Prediction Compatibility Objective}
We ablate on the heuristic rules designed for the prediction compatibility objective, 
i.e., the rendering loss implemented with 2DGS rasterization.
First, we try different scale factors $\alpha$ between the final 2DGS radius and the point map gradient magnitude, i.e.,
$\mathbf{r}_{i,x,y} = \alpha |\mathbf{n}_{i,x,y}[z]| \cdot [\|\nabla_x \mathbf{p}_{i,x,y}\|, \|\nabla_y \mathbf{p}_{i,x,y}\|]$.
We also report the results of optimizing the 2DGS parameters directly, with MVT's predictions as the initialization.
The results are shown in Tab.~\ref{tab:suppl_ablation_combined}.
First, our method is robust to a wide range of radius scale factors, from 0.05 to 5.
More specifically, smaller radius scale factor performs slightly better.
With direct 2DGS parameter optimization, the performance deteriorates significantly.
This observation justifies that our heuristic rules are necessary for the success of our method by leaving
the adaptation to the MVT's parameters.
\subsection{Fine-tuning Strategy}
We ablate on different LoRA ranks, including 1, 4 (default), and 16. 
The results are shown in Tab.~\ref{tab:suppl_ablation_combined}.
We find that our method achieves similar performance with different LoRA ranks.
At the same time, even with 1 LoRA rank, our method already outperforms the base image-only model VGGT.
\begin{table}[h]
    \centering
    \caption{
        \textbf{Ablation: Target view rendering loss and LoRA rank (ETH3D).}
        Acc., Comp., and N.C. are reported (lower is better for Acc./Comp., higher is better for N.C.).
    }
    \tablestyle{2pt}{1}
    \resizebox{1.0\columnwidth}{!}{
    \begin{tabular}{l@{\hspace{0.8em}}cc@{\hspace{0.8em}}cc@{\hspace{0.8em}}cc}
        \toprule[0.1em]
        {\textbf{Setting}} &
        \multicolumn{2}{c}{\textbf{Acc.} $\downarrow$}  &
        \multicolumn{2}{c}{\textbf{Comp.} $\downarrow$} &
        \multicolumn{2}{c}{\textbf{N.C.} $\uparrow$} \\
        \cmidrule(r){2-3} \cmidrule(r){4-5} \cmidrule(r){6-7}
         & Mean & Med. & Mean & Med. & Mean & Med. \\
        \midrule[0.08em]
        VGGT (base) & 0.280 & 0.185 & 0.305 & 0.182 & 0.853 & 0.950 \\
        \midrule[0.08em]
        \multicolumn{7}{l}{\textit{Rendering loss}} \\
        Radius scale = 0.5 (def.) & \textbf{0.114} & \textbf{0.060} & \textbf{0.116} & \textbf{0.054} & \textbf{0.901} & \textbf{0.984}  \\
        Radius scale = 0.05 & 0.122 & 0.066 & 0.124 & 0.058 & 0.895 & 0.983 \\
        Radius scale = 5        & 0.143 & 0.072 & 0.207 & 0.123 & 0.871 & 0.972 \\
        Optimize 2DGS params      & 0.895 & 0.771 & 0.809 & 0.476 & 0.565 & 0.598 \\
        \midrule[0.08em]
        \multicolumn{7}{l}{\textit{LoRA rank}} \\
        1   & 0.122 & 0.072 & 0.193 & 0.123 & 0.880 & 0.963 \\
        4 (default)  & \textbf{0.114} & \textbf{0.060} & \textbf{0.116} & \textbf{0.054} & \textbf{0.901} & \textbf{0.984} \\
        16  & 0.127 & 0.074 & 0.135 & 0.073 & 0.894 & 0.983 \\
        \bottomrule[0.1em]
    \end{tabular}
    }
    \label{tab:suppl_ablation_combined}
\end{table}

\section{Implementation Details}
For reconstruction tasks, 
we only use photometric loss to realize the prediction compatibility objective.
We set rotation loss weight $\mu_1 = 1.0$, translation loss weight $\mu_2 = 2$, and focal length loss weight $\mu_3 = 0.01$.
For ETH3D and 7-Scenes datasets, we use the a weaker photometric loss weight, i.e., $\lambda_1 = 0.2$.
For DTU and NRGBD datasets, we use a stronger photometric loss weight, i.e., $\lambda_1 = 1.0$.
All the datasets set test-time training steps as 40 except DTU adopts 50 steps due to more challenging scenes.
The learning rates are set to $5\times10^{-4}$, $1\times10^{-4}$, $2\times10^{-4}$, and $1\times10^{-3}$, for ETH3D, NRGBD, DTU, and 7-Scenes datasets, respectively.
For the camera pose estimation task, 
we only use the geometric loss to realize the prediction compatibility objective.
The test-time training steps are set to 40 and learning rate is set to $2\times10^{-4}$.

\section{Robustness to the Prior Noise}
We test the robustness of our method to camera pose and intrinsic noise.
We perturb the camera pose and intrinsic parameters by adding a small random perturbation to the ground truth values.
We report the results in Tab.~\ref{tab:suppl_noise_robustness}.
Although the performance deteriorates as the perturbation increases, 
our method still outperforms the base image-only model VGGT in most cases.
\begin{table}[h]
    \centering
    \caption{
        \textbf{Noise robustness (ETH3D).}
        We report mean Acc., mean Comp., and mean N.C. (lower is better for Acc./Comp., higher is better for N.C.).
    }
    \tablestyle{2pt}{1.0}
    \resizebox{1.0\columnwidth}{!}{
    \begin{tabular}{c@{\hspace{0.8em}}c@{\hspace{0.8em}}c@{\hspace{0.8em}}ccc}
        \toprule[0.1em]
        \textbf{Rot pert. (deg)} & \textbf{Trans pert. (\%)} & \textbf{Focal pert. (\%)} & \textbf{Acc.} $\downarrow$ & \textbf{Comp.} $\downarrow$ & \textbf{N.C.} $\uparrow$ \\
        \midrule[0.08em]
        \multicolumn{3}{c}{VGGT (base)} & 0.280 & 0.305 & 0.853 \\
        \midrule[0.08em]
        0 & 0 & 0 & 0.1765 & 0.1816 & 0.8822 \\
        \midrule[0.08em]
        1 & 0 & 0 & 0.1755 & 0.1755 & 0.8791 \\
        3 & 0 & 0 & 0.2111 & 0.2108 & 0.8669 \\
        5 & 0 & 0 & 0.2344 & 0.2143 & 0.8569 \\
        \midrule[0.08em]
        0 & 1 & 0 & 0.1702 & 0.1700 & 0.8790 \\
        0 & 5 & 0 & 0.2383 & 0.2013 & 0.8585 \\
        0 & 10 & 0 & 0.2816 & 0.2381 & 0.8489 \\
        \midrule[0.08em]
        0 & 0 & 1 & 0.1805 & 0.1657 & 0.8822 \\
        0 & 0 & 5 & 0.2328 & 0.2083 & 0.8740 \\
        0 & 0 & 10 & 0.2549 & 0.2654 & 0.8527 \\
        \midrule[0.08em]
        1 & 1 & 1 & 0.1643 & 0.1409 & 0.8850 \\
        3 & 5 & 5 & 0.2588 & 0.2112 & 0.8625 \\
        5 & 10 & 10 & 0.2901 & 0.2708 & 0.8413 \\
        \bottomrule[0.1em]
    \end{tabular}
    }
    \label{tab:suppl_noise_robustness}
\end{table}

\begin{figure*}[t]
    \centering
    \includegraphics[width=1.0\textwidth]{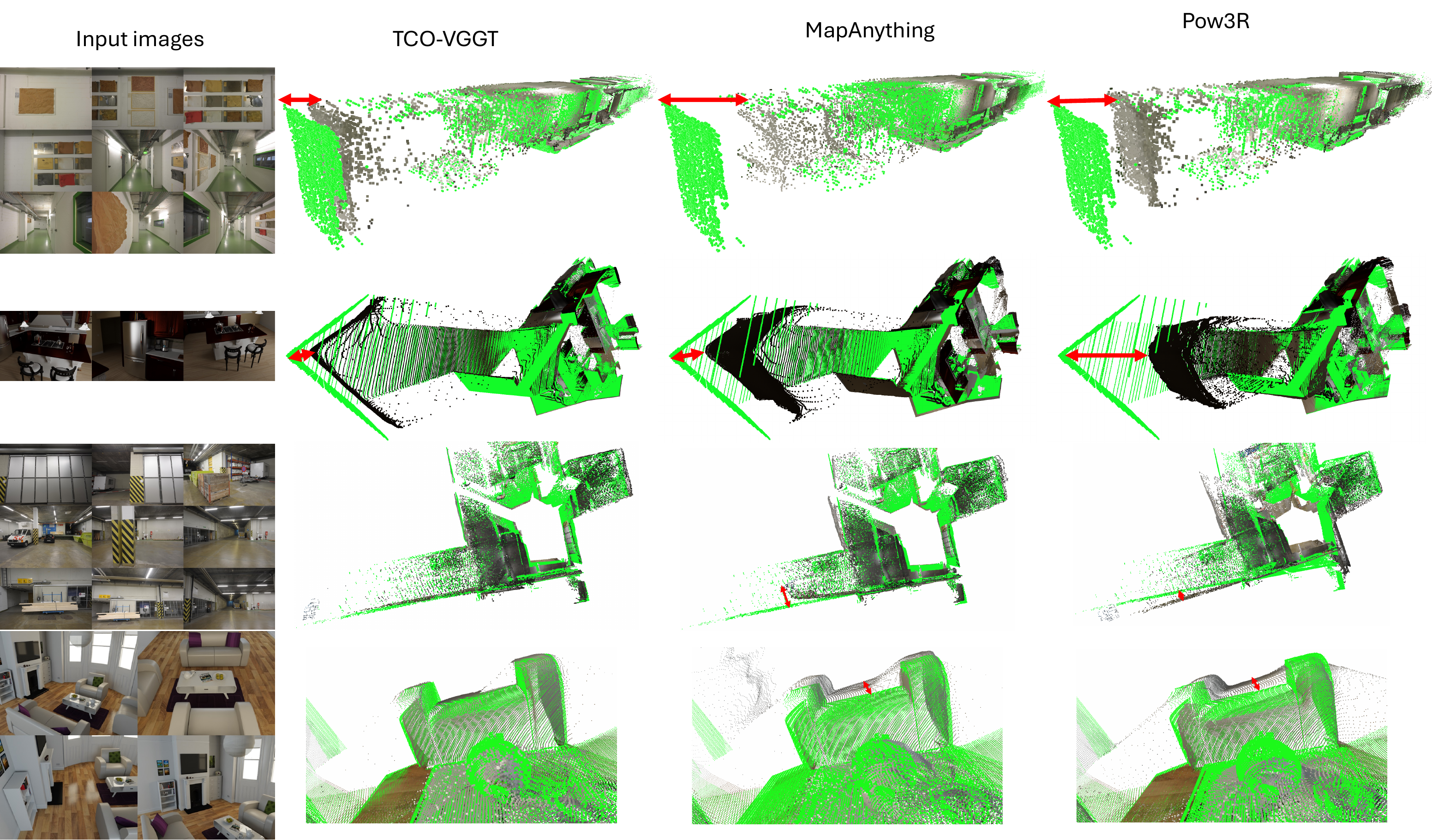}
    \caption{
        \textbf{Fine-grained Qualitative Results.} We compare TCO-VGGT with prior-aware feed-forward methods, including Pow3R and MapAnything.
        In each grid cell, the predicted geometry is overlaid with the ground truth geometry, whose points are shown in \textcolor{green}{green}.
        Discrepancies between the predicted and ground-truth geometries are highlighted by \textcolor{red}{red} double arrows, whose lengths indicate the magnitude of the errors.
        TCO-VGGT exhibits much smaller discrepancies than MapAnything and Pow3R in both scene structure and boundary regions.
    }
    \label{fig:fine_qualitative}
\end{figure*}

\section{Test-time Inference Time and Limitations}
In this section, 
we report the test-time inference time of our method on the ETH3D dataset
under different settings.
As shown in Tab.~\ref{tab:suppl_inference_time_stats},
inference time is a limitation of our method.
By trading off some efficiency, 
we improve the performance of our method on a series of benchmarks.
We will leave the exploration of more efficient test-time methods for future work.
\begin{table}[h]
    \centering
    \caption{
        \textbf{Test-time training steps vs. time (10 views).}
        The reported times are wall-clock seconds per scene. The image resolution is 392(H)$\times$518(W).
        The base model is VGGT. The GPU is NVIDIA TITAN RTX.
    }
    \tablestyle{1.5pt}{0.95}
    \resizebox{1\columnwidth}{!}{
    \begin{tabular}{c@{\hspace{0.8em}}c@{\hspace{0.8em}}c@{\hspace{0.8em}}c@{\hspace{0.8em}}c@{\hspace{0.8em}}c}
        \toprule[0.1em]
        \textbf{TTT Steps} & \textbf{Render Loss} & \textbf{Avg (s)} & \textbf{Std (s)} & \textbf{Min (s)} & \textbf{Max (s)} \\
        \midrule[0.08em]
        0  & \xmark     & 1.301  & 0.198 & 1.104  & 1.499 \\
        10 & \xmark     & 43.630 & 0.218 & 43.412 & 43.848 \\
        10 & \checkmark & 45.598 & 0.440 & 45.158 & 46.037 \\
        40 & \checkmark & 161.547 & 1.424 & 160.123 & 162.971 \\
        \bottomrule[0.1em]
    \end{tabular}
    }
    \label{tab:suppl_inference_time_stats}
\end{table}
\begin{figure*}[t]
    \centering
    \includegraphics[width=1.0\textwidth]{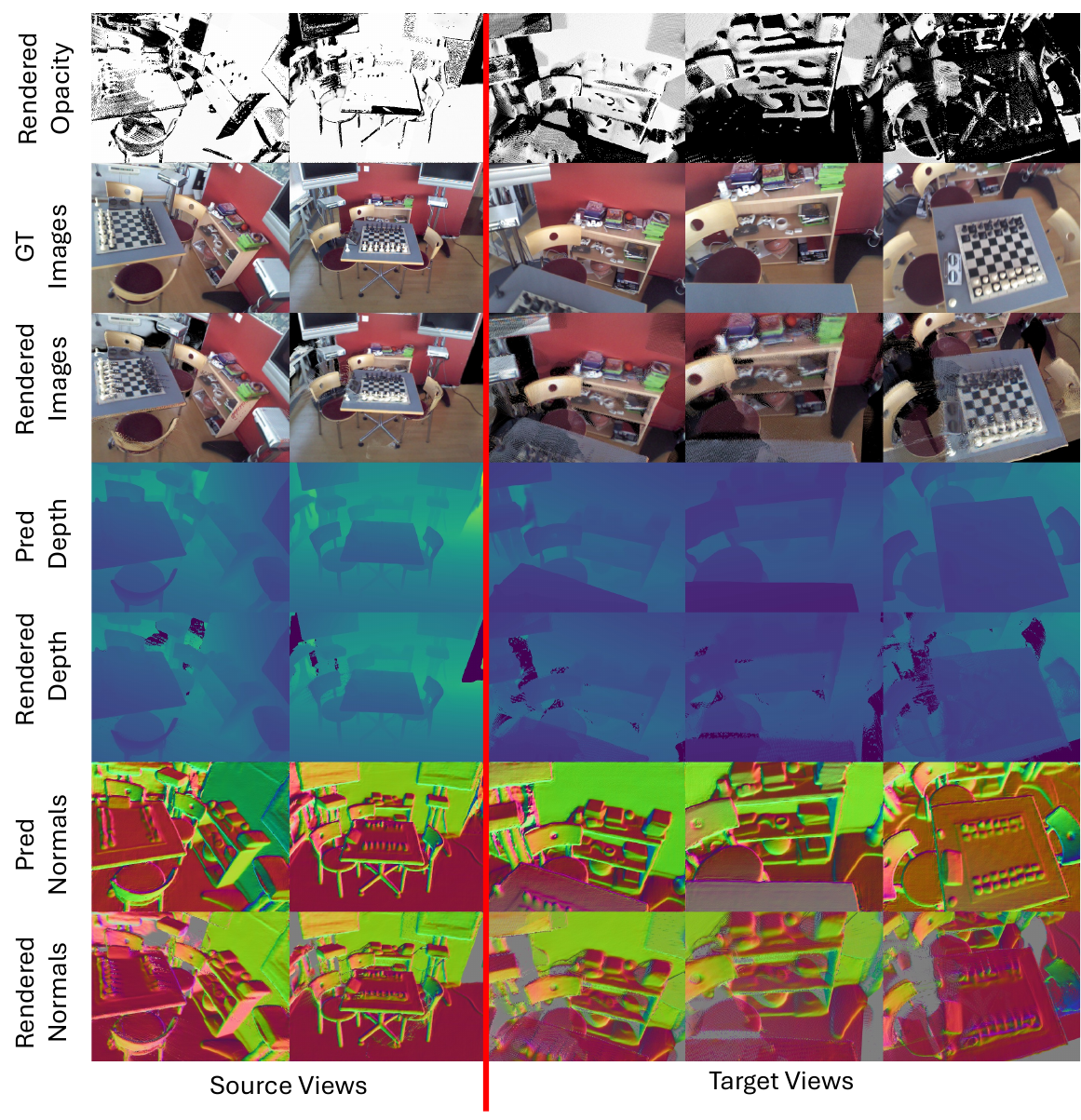}
    \caption{
        \textbf{2DGS Rendering Visualization.} We visualize the 2DGS rendering process for one scene from 7-Scenes.
        As shown in the Rendered Image row, our 2DGS heuristic parameterization produces rendered images that closely match the ground-truth images.
        We also compare the depth maps and normal maps rendered from 2DGS with the corresponding ground-truth depth and normal maps, i.e., those directly predicted from the MVT views.
    }
    \label{fig:2dgs_rendering}
\end{figure*}